\documentclass[10pt,twocolumn,letterpaper]{article}

\usepackage{iccv}
\usepackage{times}
\usepackage{epsfig}
\usepackage{graphicx}
\usepackage{amsmath}
\usepackage{amssymb}

\usepackage{amsfonts}
\usepackage{enumitem}
\usepackage{multirow}
\usepackage{booktabs} 
\usepackage{comment}
\usepackage{subfig}

\newcommand{\tablestyle}[2]{\setlength{\tabcolsep}{#1}\renewcommand{\arraystretch}{#2}\centering\footnotesize}

\newcommand{\ieno}{\textit{i}.\textit{e}.}
\newcommand{\tcb}{\textcolor{black}}
\newcommand{\tcr}{\textcolor{black}}

\newcommand{\tcp}{\textcolor{black}}
\newcommand{\tcg}{\textcolor{black}} 

\usepackage[pagebackref=true,breaklinks=true,letterpaper=true,colorlinks,bookmarks=false]{hyperref}

\iccvfinalcopy 

\def\iccvPaperID{3055} 

\ificcvfinal\fi

\begin{document}


\title{Re-energizing Domain Discriminator with Sample Relabeling \\ for Adversarial Domain Adaptation}


\author{{Xin Jin{$^{1}$}} \qquad Cuiling Lan{$^{2}$}\thanks{Corresponding Author.} \qquad   Wenjun Zeng{$^{2}$} \qquad  Zhibo Chen{$^{1*}$}\\
	\normalsize
	$^{1}$\	University of Science and Technology of China ~~ $^{2}$\,Microsoft Research Asia, Beijing, China\\
	\normalsize
	{\tt\small jinxustc@mail.ustc.edu.cn\quad \{culan,wezeng\}@microsoft.com\quad chenzhibo@ustc.edu.cn}
	}

\maketitle
\ificcvfinal\thispagestyle{empty}\fi

\begin{abstract}
   Many unsupervised domain adaptation (UDA) methods exploit domain adversarial training to align the features to reduce domain gap, where a feature extractor is trained to fool a domain discriminator in order to have aligned feature distributions. 
   The discrimination capability of the domain classifier w.r.t.~the increasingly aligned feature distributions deteriorates as training goes on, thus cannot effectively further drive the training of feature extractor. 
   In this work, we propose an efficient optimization strategy named Re-enforceable Adversarial Domain Adaptation (RADA) which aims to re-energize the domain discriminator during the training by \tcr{using dynamic domain labels.} Particularly, we relabel the well aligned target domain samples as source domain samples on the fly. 
   Such relabeling makes the less separable distributions more separable, and thus leads to a more powerful domain classifier w.r.t.~the new data distributions, which in turn further drives feature alignment. Extensive experiments on multiple UDA benchmarks demonstrate the effectiveness and superiority of our RADA. 
   

\end{abstract}

\begin{figure}
  \centerline{\includegraphics[width=1.0\linewidth]{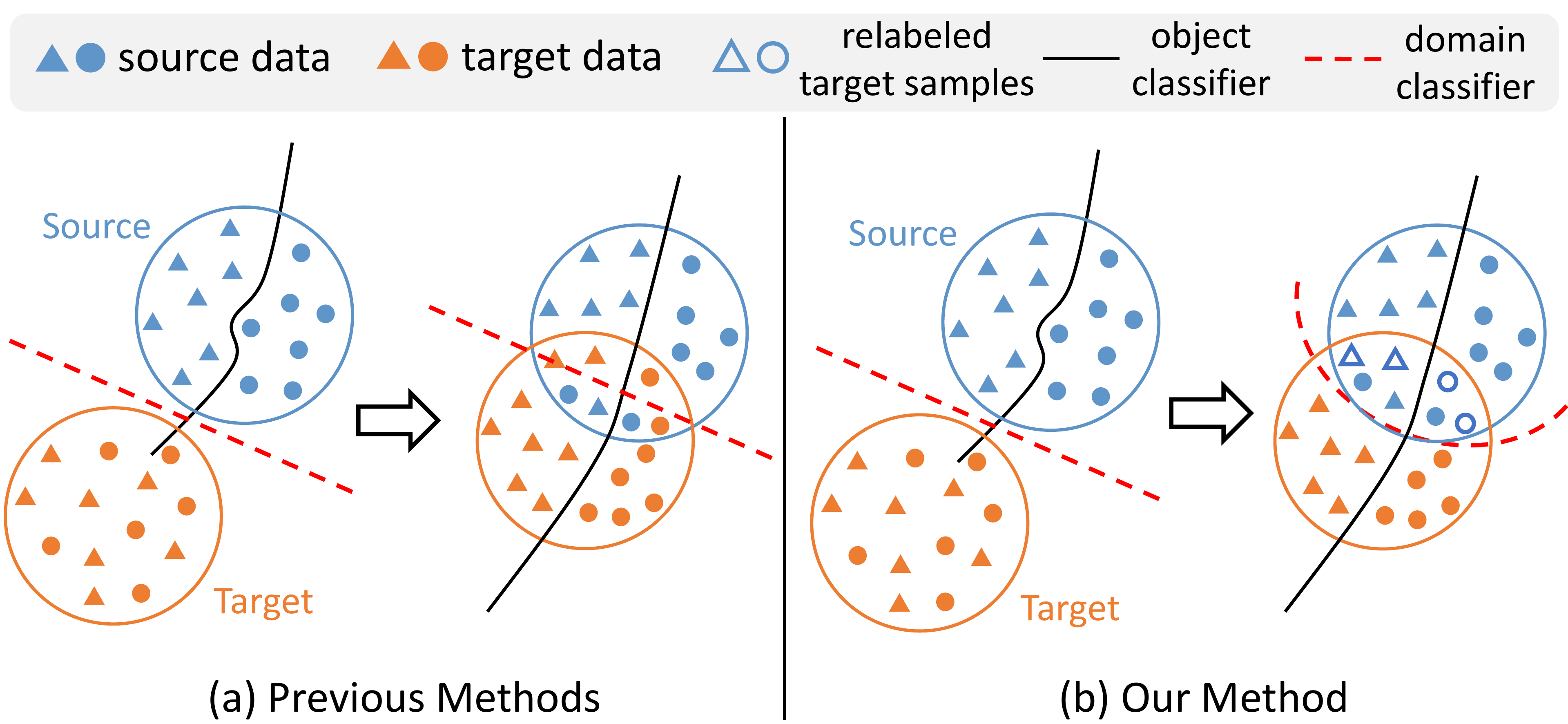}}
  \vspace{-2mm}
  \caption{Motivation and our idea. (a) Previous methods align domain distributions by adversarially training the domain classifier with static domain labels of the samples. The discrimination capability of the domain classifier w.r.t. the increasingly aligned feature distributions deteriorates as training goes on (\ieno, more samples are inseparable), which in turn provides less driving power to the feature extractor for alignment and prevents effective optimization. (b) We propose a new solution which allows dynamic domain labels. We relabel the ``well aligned" target samples as source domain, which makes the two less separable distributions more separable and thus leads to a more powerful domain classifier, which in turn further drives feature alignment.}
  \vspace{-6mm}
  \label{fig:motivation}
\end{figure}

\vspace{-3mm}
\section{Introduction}
The rapid development of deep learning has helped achieve significant improvements for many computer vision tasks. Those learned models usually have high performance on the test data that share similar characteristics as the training data, but suffer from significant performance degradation when deployed in new environments~\cite{ganin2014unsupervised,ganin2016domain,jin2020style}. There are \emph{domain gaps} between the training source data (source domain) and the testing target data (target domain). Besides, annotating the data for the target domain is expensive and time-consuming. To mitigate the notorious domain gap issue without manual annotation efforts, unsupervised domain adaptation (UDA)~\cite{ganin2014unsupervised,ganin2016domain,haeusser2017associative,hu2018duplex,zhuo2019unsupervised,kang2019contrastive,cui2020nnm} is recently extensively explored, which aims to learn from labeled source domain and unlabeled target domain.

For domain adaptation, theoretical analysis by Ben-David~\etal~\cite{ben2007analysis} shows that reducing the feature differences between the source and target domains can reduce the upper bound of the target domain error. 
Many domain adaptation methods tend to learn domain invariant/transferable feature representations~\cite{gong2012geodesic,gretton2012kernel,gong2013connecting,ni2013subspace}. Inspired by the Generative Adversarial Networks (GANs)~\cite{goodfellow2014generative}, adversarial learning has been successfully applied for UDA~\cite{ganin2016domain,tzeng2017adversarial,chen2018re,sankaranarayanan2018generate,volpi2018adversarial,long2018conditional,zhang2019domain,cui2020gradually}. The core idea of adversarial domain adaptation approaches is to train a domain discriminator/classifier to distinguish between the source and target domains and train the generator/feature extractor to minimize the feature discrepancy between the source and target domain in order to fool the discriminator in a minmax two-player game. High discrimination capability for the domain discriminator is desired in order to be able to drive the feature alignment.
Typically, a domain adversarial neural network (DANN)~\cite{ganin2016domain} introduces a gradient reversal layer (GRL) for adversarial training, where the ordinary gradient descent is applied for optimizing the domain classifier and the sign of the gradient is reversed when passing through the GRL to optimize the feature extractor. 
Conditional domain adversarial network (CDAN)~\cite{long2018conditional} improves DANN by conditioning the domain discriminator on both the object classification predictions/likelihoods and the extracted features. 

All these methods train the domain discriminator with the static domain labels of the source and target samples. However, as shown in Figure~\ref{fig:motivation} (a), as the adversarial training goes on, the feature distributions for the source domain and target domain are increasingly aligned. \textbf{The discrimination capability of the domain discriminator/classifier w.r.t. the more aligned distributions is weaker than that w.r.t. the earlier less aligned distributions. Such degradation of discrimination capability in turn provides less driving power to the feature extractor for alignment, hindering effective optimization}, even though there are still not aligned samples in the feature space. 

\begin{figure}
  \centerline{\includegraphics[width=1.0\linewidth]{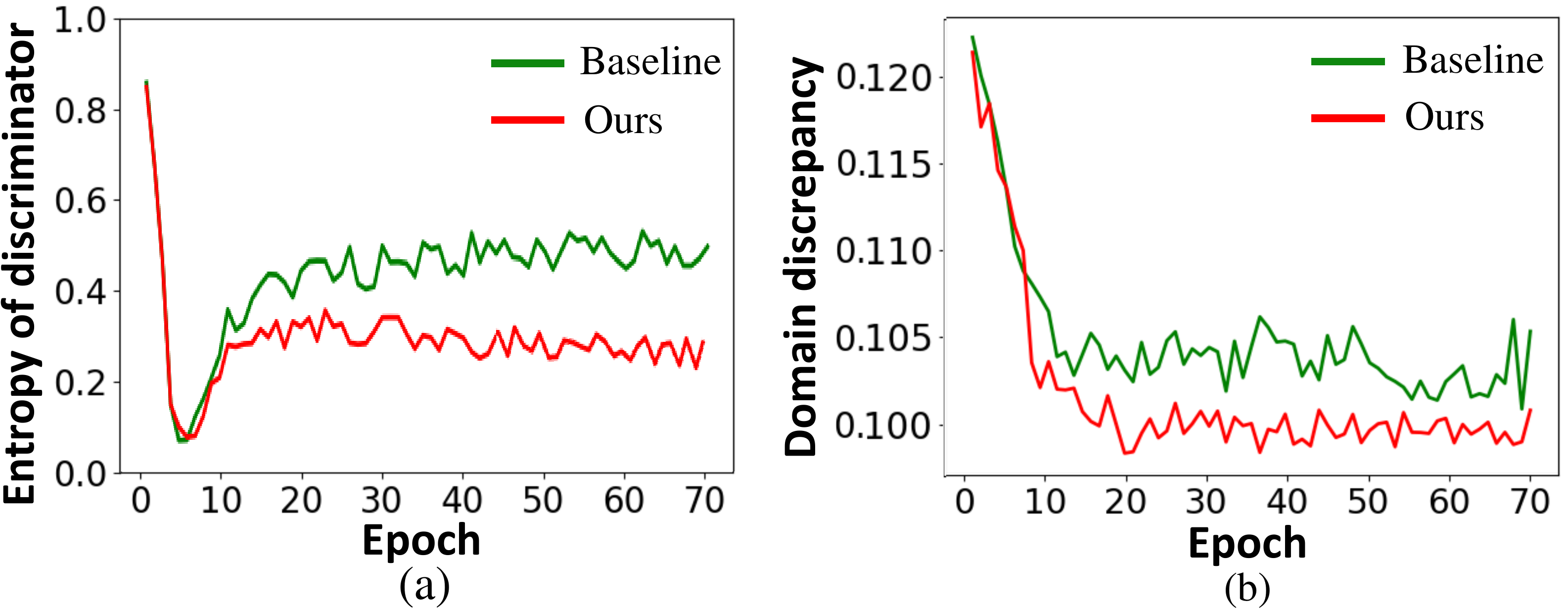}}
  \vspace{-3mm}
  \caption{The variation of (a) the discrimination capability of the domain discriminator (measured by entropy of domain classification) and (b) alignment state (measured by domain discrepancy measure of MMD) in the training. For the baseline scheme CDAN~\cite{long2018conditional} (marked by green), the discrimination capability of the domain discriminator deteriorates w.r.t.~the gradually aligned distributions after the initial dip of the entropy, which in turn provides less driving power to the feature extractor for alignment. In contrast, thanks to our strategy of re-labeling the aligned target samples as source samples, our scheme (marked by red) could improve the discrimination capability of the domain discriminator (\ieno, preventing the increasing of the entropy) and thus in turn further drives feature alignment. These experiments are conducted on Office-31 of the setting W$\rightarrow$A. \tcr{Note that more analysis can be found in our \textbf{Supplementary}.}}
  
  \vspace{-5mm}
  \label{fig:motivation2}
\end{figure}

To better understand what is happening, on top of a representative baseline scheme CDAN~\cite{long2018conditional}, we observe the variation of the discrimination capability of the domain discriminator by calculating the average entropy of domain classification for all the training samples at each training epoch (see  Figure~\ref{fig:motivation2}~(a)), where an epoch is a single pass through the full training set. As we know, a larger entropy of domain classification (\ieno, the discriminator has larger ambiguity on the domain classes of the samples) indicates a poorer discrimination capability of the domain classifier. We also observe how well the alignment is by calculating a domain discrepancy measurement Maximum Mean Discrepancy~(MMD) \cite{long2015learning,borgwardt2006integratingmmd,yan2017mind} (see Figure~\ref{fig:motivation2}~(b)).

Figure~\ref{fig:motivation2}~(a) reveals that the average entropy of the baseline scheme goes through a process of fast decreasing and then slowly rises with fluctuation. 1) In the early epochs, the successive training of domain classifier increases its discrimination power and thus the entropy decreases.
2) Meanwhile, as the training goes on, the feature distributions for the source domain and target domain are increasingly aligned, \ieno, the domain discrepancy decreases as shown in Figure~\ref{fig:motivation2}~(b). This could adversarially increase the entropy. 
3) The optimization progresses/paces of the feature extractor and the domain discriminator are usually not the same. Around 5 to 15 epochs, the domain discrepancy continues to decrease even when the discrimination capability reduces (\ieno, entropy increases). This may be because the optimization of feature extractor lags behind that of the domain discriminator and thus the feature extractor can be further optimized. 4) The discrimination capability of the domain classifier w.r.t.~the more aligned distributions is becoming weaker than that w.r.t.~the earlier less aligned distributions (the entropy increases). The persistent degradation  
in turn provides less driving power to the feature extractor for alignment, hindering effective optimization (where the alignment state cannot be effectively improved after 15 epochs, please refer to the green curve in Figure~\ref{fig:motivation2}~(b)). \textbf{Thus, a mechanism that could enhance/improve on the fly the discrimination capability of the domain classifier to re-energize the adversarial training is highly desired.}


Motivated by this, in this paper, we propose a new optimization strategy named Re-enforceable Adversarial Domain Adaptation (RADA) which aims to re-energize the domain discriminator/classifier during the training. Particularly, instead of training the domain classifier using static domain labels for the source and target samples (see Figure \ref{fig:motivation}~(a)), we propose to  \tcr{exploit dynamic domain labels}, where we relabel the ``well aligned" target samples, \ieno, those close to source domain, as source domain samples, \tcr{in each mini-batch to be optimized/trained}. As illustrated in Figure \ref{fig:motivation}~(b), this makes the two less separable distributions more separable and thus leads to a more powerful domain discriminator. 
Such re-energized domain discriminator in turn further drives the feature alignment for the feature extractor.
From Figure~\ref{fig:motivation2}, we can observe that after using our proposed data relabeling strategy, the domain classification entropy is much lower than that of the baseline scheme and the domain discrepancy is also smaller.  

We summarize our main contributions as follows:
\begin{itemize}[leftmargin=*,noitemsep,nolistsep]

\item We pinpoint that the popular adversarial domain adaptation approaches in general face optimization difficulty, which is caused by the deteriorated discrimination capability of the domain classifier as the feature distributions become increasingly aligned in training.  

\item To alleviate the deterioration of the discrimination capability of the domain discriminator, we propose an efficient optimization strategy named Re-enforceable Adversarial Domain Adaptation (RADA), which is capable of re-energizing the domain discriminator during training which in turn further drives feature alignment. We achieve this by relabeling the well aligned target samples as source domain samples for online training of the domain discriminator.  

\end{itemize}

We will demonstrate the effectiveness of the proposed RADA on top of multiple widely-used adversarial learning based domain adaptation baselines. RADA significantly outperforms the state-of-the-art UDA approaches. RADA is simple yet effective and can be used as a \emph{plug-and-play} optimization strategy for existing adversarial learning based UDA approaches. \tcb{Note that we do not make any change to the network architecture of the domain discriminator, which makes the RADA friendly to many adversarial domain adaptation methods.}  
We will release our code upon acceptance.




\section{Related Work}

\noindent\textbf{Unsupervised Domain Adaptation (UDA).}
UDA has achieved great progress with the advancement of deep learning~\cite{ganin2014unsupervised,ganin2016domain,haeusser2017associative,hu2018duplex,kang2019contrastive,cui2020nnm}. 
These UDA methods could be mainly categorized into three classes: pixel-level translation based methods~\cite{bousmalis2016domain,bousmalis2017unsupervised,liu2017unsupervised}, explicit domain alignment based methods~\cite{gretton2012kernel,long2015learning,long2017deep,sun2016deep,yan2017mind}, and adversarial learning based methods~\cite{ganin2016domain,tzeng2017adversarial,chen2018re,sankaranarayanan2018generate,volpi2018adversarial,long2018conditional,zhang2019domain,cui2020gradually}. \tcb{The third category predominates in recent years due to their superior performance and we will focus and elaborate on them}. 

Inspired by Generative adversarial networks (GANs) \cite{goodfellow2014generative}, adversarial domain adaptation has been widely explored~\cite{chen2018re,sankaranarayanan2018generate,volpi2018adversarial}. 
The core idea of adversarial UDA approaches is to train a domain discriminator to distinguish the source and target domains and train a feature extractor to learn domain-invariant
features to fool the discriminator. Domain Adversarial Neural Network (DANN)~\cite{ganin2016domain} is a representative work, where a domain classifier is connected to the feature extractor via a gradient reversal layer (GRL). Subsequently, many well-designed adversarial UDA variants are proposed, such as ADDA \cite{tzeng2017adversarial}, CyCADA \cite{hoffman2018cycada}, SBADA~\cite{russo2018source}, CDAN~\cite{long2018conditional}, MCD~\cite{saito2018maximum}, MSTN~\cite{xie2018learning} and TADA~\cite{wang2019transferable}. Symnets~\cite{zhang2019domain} builds a symmetric design of source and target task classifiers, on which the domain discrimination and domain confusion training are based. TADA~\cite{wang2019transferable} is also built upon adversarial domain adaptation framework, which adds adversarial alignment constraints on both of transferable local regions and global images through two local/global attention modules.
\tcb{MSTN~\cite{xie2018learning} improves the adversarial UDA framework by additionally leveraging clustering technique to align labeled source centroid and pseudo-labeled target centroid, where features in the same class but different domains are expected to be mapped to nearby. Similarly, PFAN~\cite{chen2019progressive} performs feature alignment between category-wise prototypes, with the number of target domain samples progressively increasing.}
CDAN~\cite{long2018conditional} proposes to condition the domain discriminator on the discriminative information conveyed in the classifier predictions (class likelihood). Besides, for safe transfer, it prioritizes over those easy-to-transfer samples with {confident predictions of object classes (measured by the entropy)} by reweighting each sample by an entropy-aware weight. 
Recently, Cui~\etal~\cite{cui2020gradually} equip the adversarial adaptation framework with a Gradually Vanishing Bridge (GVB) mechanism, which reduces the transfer difficulty by gradually reducing the domain-specific characteristics in domain-invariant representations. CMSS~\cite{yang2020curriculum} introduces an agent to learn a dynamic curriculum to softly select/re-weight source samples that are best suited for aligning to the target domain. 


As the optimization of the minmax game in domain adversarial adaptation proceeds, the increasingly aligned feature distributions would degrade the discrimination capability of the domain classifier, which reduces the driving power for the feature alignment. In this paper, we propose to relabel the well aligned target samples as source samples to make the less separable distributions more separable, re-energizing the entire adversarial optimization.

\noindent\textbf{Mixup Technique.}
Mixup, which mixes two samples and their labels to generate a new sample, as a data augmentation/regularization technique, was first introduced in~\cite{zhang2018mixup} and was widely applied in deep learning pipelines to improve performance~\cite{he2019bag,cubuk2019autoaugment,yun2019cutmix,guo2019mixup}. The study of using Mixup for domain adaptation is rare~\cite{xu2020adversarial,wu2020dual} and remains an open problem. Xu~\etal~\cite{xu2020adversarial} present adversarial domain adaptation with domain mixup (DM-ADA) to guarantee domain-invariance in a more continuous latent space, where two samples (one from source domain and the other from target domain) and their domain labels are softly mixed to explore inter-domain information. Wu~\etal~\cite{wu2020dual} propose a dual mixup regularized learning (DMRL) method for UDA, which jointly conducts category and domain mixup regularization on the pixel level to guide the object classifier in enhancing consistent predictions in-between samples and enrich the intrinsic structures of the latent space. These mixup methods aim to explore inter-domain information. \tcb{In contrast}, we aim to enhance the continuity within the source domain after the re-partition of domains, where the samples to be mixed share the same source domain labels.

\section{Re-enforceable Adversarial Domain Adaptation (RADA)}

\begin{figure*}
  \centerline{\includegraphics[width=1.0\linewidth]{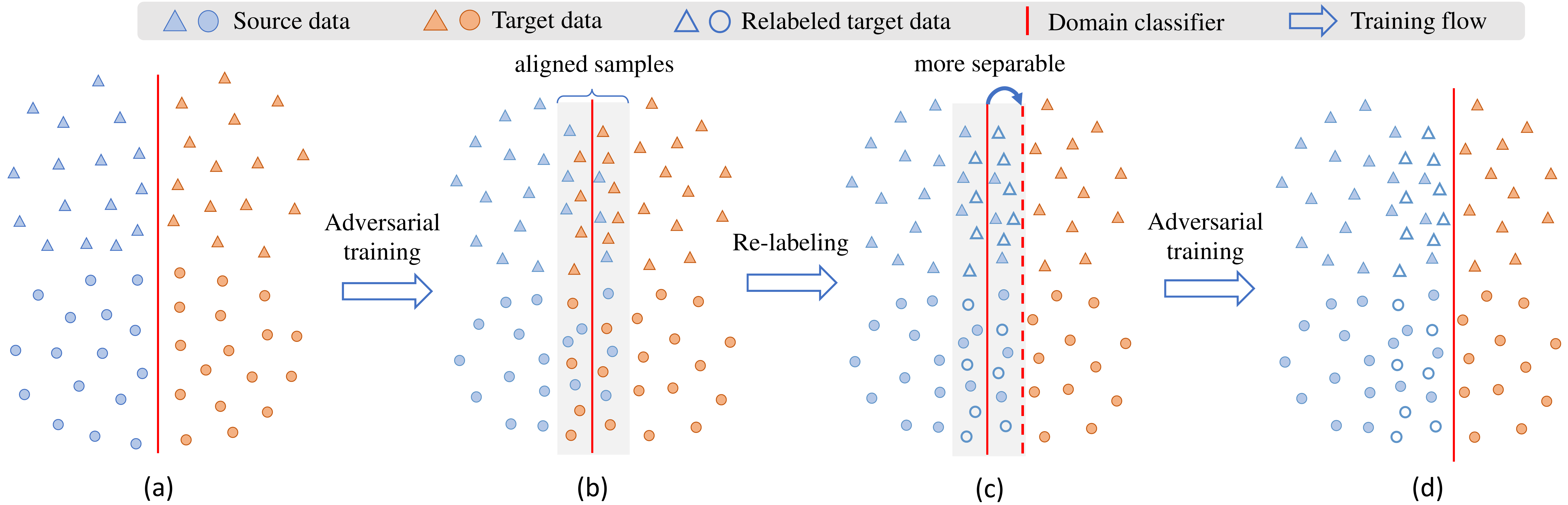}}
  \vspace{-2mm}
  \caption{Illustration of our Re-enforceable Adversarial Domain Adaptation (RADA) optimization using toy examples. (a) Initial feature distributions of source and target data sets. (b) As the online adversarial training goes on, the source and target  distributions are increasingly aligned. Some samples are already aligned such that the domain classifier cannot distinguish them well while some other samples are still not aligned. The ambiguous samples deteriorate the discrimination capability of the domain classifier. (c) We relabel the well aligned target samples as source domain samples, which makes the less separable distributions more separable. With this strategy, we have a re-enforceable domain classifier, which further re-energizes the domain adversarial optimization in (d).}
  
  
  \label{fig:pipeline}
  \vspace{-4mm}
\end{figure*}


We propose an efficient domain adaptation optimization strfategy, named Re-enforceable Adversarial Domain Adaptation (RADA), which re-energizes the domain classifier/discriminator during the adversarial training \tcb{by using sample domain re-labeling strategy}. Figure~\ref{fig:pipeline} illustrates the main procedure of our RADA optimization. As the adversarial training goes on, the source and target distributions are increasingly aligned. As illustrated in Figure~\ref{fig:pipeline}~(b), some samples are already aligned such that the domain classifier cannot distinguish them well while some other samples are still not aligned. \tcb{As revealed in Figure~\ref{fig:motivation2},} the discrimination capability of the domain classifier (given a fixed model size) \tcb{w.r.t. the more aligned distributions} deteriorates since the aligned samples are inseparable, which in turn results in less driving power for the feature alignment. 

We re-energize \tcb{this} domain discriminator by relabeling the well-aligned target domain samples as source domain, leading to a clear boundary for the two new distributions. Particularly, as illustrated in Figure~\ref{fig:pipeline}~(c), we relabel the well aligned target samples as source domain samples and train the domain classifier under the updated partition. 
\tcb{To determine which target samples are to be relabeled in the training, we use} a metric to evaluate how well a target sample is \tcb{aligned to the source domain, by simply taking into account the certainty~(entropy) of the prediction of the domain classifier. When the entropy is large, it indicates the sample is close to the domain classification boundary and we will relabel it to the source domain.} 

For unsupervised domain adaption classification, we denote the source domain as $\mathcal{D}_S=\{(\mathbf{x}_i^s,y_i^s) \}_{i=1}^{N_s}$ with $N_s$ labeled samples covering $C$ classes, and the target domain as $\mathcal{D}_T=\{\mathbf{x}_j^t \}_{j=1}^{N_t}$ with $N_t$ unlabeled samples that belong to the same $C$ classes. UDA aims to train on $\mathcal{D}_S$ and $\mathcal{D}_T$ to achieve high accuracy on the target domain test set.

To be self-contained, we first review the adversarial domain adaptation framework in Section \ref{sec:ADA_baseline}. On top of the framework, we introduce our proposed training strategy: Re-enforceable Adversarial Domain Adaptation (RADA) in Section \ref{subsec:RADA}.

\subsection{Recap of Adversarial Domain Adaptation}
\label{sec:ADA_baseline}

Adversarial Domain Adaptation aims to learn domain invariant representations for UDA, by adversarially training the feature extractor and domain discriminator/classifier. The adversarial domain adaptation~\cite{ganin2016domain,long2018conditional,zhang2019domain,cui2020gradually} can be achieved by optimizing object classification loss $\mathcal{L}_{cls}$ and domain adversarial loss $\mathcal{L}_{adv}$ ~\cite{ganin2015unsupervised,hoffman2018cycada,long2018conditional}:
\begin{align}
         \mathcal{L}_{cls} = &\frac{1}{N_s} \sum_{i=1}^{N_s} \mathcal{L}_{ce} (C(F({x}_i^s))),y_i^s), \\
        \mathcal{L}_{adv}= &-\frac{1}{N_s} \sum_{i=1}^{N_s} \log D(F({x}_i^s)) \nonumber\\
    	&- \frac{1}{N_t}\sum_{j=1}^{N_t} \log (1- D(F({x}_j^t))),
        \label{eq:loss} 
\end{align}
where $F, C, D$ denote the feature extractor, the object classifier, and domain discriminator/classifier, respectively. $\mathcal{L}_{ce}$ is the cross-entropy loss. The total training objective is described as follows:
\vspace{-2mm}
\begin{align}
        &\min_D \mathcal{L}_{adv},
        \label{eq:advopt-1}  \\
        &\min_{F, C} \mathcal{L}_{cls} - \lambda \mathcal{L}_{adv},
        \label{eq:advopt-2} 
\end{align}
where the domain classifier $D$ is trained to discriminate the domains. Conversely, the feature extractor is trained to extract features that make it difficult for the domain classifier to discriminate the domains, resulting in domain invariant features. $\lambda$ denotes a hyper-parameter for balancing the losses. 
\tcg{Gradient reversal layer (GRL)~\cite{ganin2015unsupervised} which connects feature extractor $F$ and the domain discriminator $D$ is usually used to achieve the adversarial function by multiplying the gradient from $D$ by a certain negative constant during the back-propagation to the feature extractor.}

\subsection{Proposed Training Strategy RADA}

\label{subsec:RADA}


During the training, the discrimination capability of the domain classifier w.r.t.~to the increasingly aligned distributions \emph{deteriorates}. This in turn negatively influences the effectiveness of the optimization towards feature alignment. 
Thus, with the adversarial optimization objectives as in (\ref{eq:advopt-1}) and (\ref{eq:advopt-2}), the networks converge to states that are less than optimal. Even though the distributions are not satisfactorily/fully aligned, the deteriorated discrimination capability of the domain classifier cannot further efficiently drive the feature alignment. We propose to re-energize the domain classifier by making the two distributions more separable. As illustrated in Figure~\ref{fig:pipeline}~(c), we relabel the ``well aligned" (ambiguous w.r.t. domain classifier) target domain samples as source domain samples. \emph{Such relabeling makes the less separable distributions more separable and thus leads to a more powerful domain classifier, which in turn further drives feature alignment.} \tcb{As revealed in Figure~\ref{fig:motivation2}~(a), the discrimination capability of the domain discriminator after using our relabeling strategy is stronger (\ieno, with smaller entropy) than the baseline.}



\noindent\textbf{Measurement of Alignment for a Target Sample.} 
To select and relabel the ``well aligned" target samples, we need to evaluate whether a target sample is ``well aligned" or not. 
As we know, when a sample is more aligned, it is harder for the domain classifier to identify its domain and has a higher uncertainty w.r.t.~to the predicted domain label of this sample. Correspondingly, the entropy of domain classification of this sample is in general higher. Therefore, we measure how well a target sample is aligned with the source domain by simply using the entropy of domain classification, 
\begin{equation}
    {\mathcal{H}}(\mathbf{p})=- p_0 \log p_0 - (1-p_0) \log (1-p_0),
\end{equation}
where \tcp{$\mathbf{p}$ = [$p_0$, $1-p_0$], with $p_0$ and $1-p_0$ denoting} the probability of predicting a sample as target domain and source domain, respectively. The larger the entropy value, the larger the ambiguity/uncertainty of the discriminator has when identifying which domain it belongs to, and the more aligned the sample is. Note that there may exist other more accurate or advanced measurement metrics, but this is not the focus of this paper and we use this simple one here.

\noindent\textbf{Well aligned Target Sample Selection and Relabeling.} \tcb{For a given target sample, when it cannot be well distinguished by the domain classifier, it could be considered approximately aligned. In other words, when the entropy of the domain classifier prediction is larger than a threshould $\tau$, we define it as a ``well aligned" sample, \tcr{which is to be relabeled as source domain}. $\tau$ is a hyper-parameter that controls the strictness of the definition of ``well align". We will study its influence in the experiments in Section \ref{sec:experiments}.}

\noindent\textbf{When to start RADA.}
\tcb{Intuitively, we could start our relabeling strategy to re-energize the domain discriminator whenever its discrimination capability begins to deteriorate. In general, at the early training stage, the optimization of the domain discriminator persistently improves its discrimination capability as revealed in Figure~\ref{fig:motivation2}~(a), where the entropy decreases rapidly (\tcr{in the first 5 epochs}). It is thus unnecessary to enable the re-energizing strategy. Inspired by the popular learning rate (lr) adjustment algorithms~\cite{prakash2019repr,bottou2012stochastic} which adjust the learning rate if no improvement is seen for a `patience' number of epochs (where `patience' is usually set to 2 to 10) \footnote{https://pytorch.org/docs/stable/optim.html?highlight=reducelronplateau\\ \#torch.optim.lr\_scheduler.ReduceLROnPlateau}, we start RADA if no improvement of the discrimination capability is seen for \tcr{a `patience' number of epochs and we denote this hyper-parameter as $K$}. 
} 

\noindent\tcb{\noindent\textbf{Adversarial Training.}
As is done in previous adversarial domain adaptation methods, we perform mini-batch level optimization where a batch consists of both source and target domain samples. Once RADA is activated, for each mini-batch, we first check whether each target sample should be relabeled as a source sample and relabel them if deemed so. Then the adversarial training is performed under the updated domain labels. 
The domain classifier is updated by minimizing the domain classification loss (\ieno, adversarial loss as in (\ref{eq:loss})) \emph{over the updated source set and target set}. Simultaneously, the feature extractor is trained to fool the domain classifier. Note that the domain relabeling has no impact on object classification loss.}

\noindent\textbf{Mixup within Updated Source Set.} To promote the continuity of feature space in the updated source set, where there may be low density space between the newly absorbed samples and previous source samples, we leverage the mixup technique ~\cite{zhang2018mixup} to softly mix the features between the previous source samples and the newly absorbed samples. 
\tcr{In a mini-batch,} we randomly select a sample from the \tcr{original} source set and a sample from the relabeled \tcr{source} sample set, where we denote their features as $\mathbf{f}_i^s$ and $\mathbf{f}_j^t$, respectively. Such mixup generates a new feature $\widetilde{\mathbf{f}^s}$ with domain label being source domain: 
\begin{equation}
    \begin{aligned}
        \widetilde{\mathbf{f}^s} &= \mathcal{M}_{\alpha}(\mathbf{f}_i^s,\mathbf{f}_j^t)=\alpha \mathbf{f}_i^s+(1-\alpha) \mathbf{f}_j^t,\\
    \end{aligned}
    \label{eq:mix}
\end{equation}
which is used to train the domain classifier and adversarially train the feature extractor. $\alpha$ is a value randomly selected from a uniform distribution, \ieno, $\alpha \sim U(0,1)$.

\section{Experiment}
\label{sec:experiments}


\subsection{Datasets and Implementation Details}~\label{sec:datasets_and_details}
\vspace{-4mm}

\noindent\textbf{Datasets.} 1) \noindent\textbf{Office-31}~\cite{saenko2010adapting} is the most widely used dataset for visual domain adaptation (DA), with 4,652 images and 31 categories collected from three distinct domains: {Amazon} (\textbf{A}), {Webcam} (\textbf{W}) and {DSLR} (\textbf{D}). We evaluate all methods on six transfer tasks \textbf{A} $\rightarrow$ \textbf{W}, \textbf{D} $\rightarrow$ \textbf{W}, \textbf{W} $\rightarrow$ \textbf{D}, \textbf{A} $\rightarrow$ \textbf{D}, \textbf{D} $\rightarrow$ \textbf{A}, and \textbf{W} $\rightarrow$ \textbf{A}, respectively.
2) \noindent\textbf{Office-Home}~\cite{venkateswara2017deep} is a more difficult dataset (with relative large domain discrepancy) than \textit{Office-31}. It consists of 15,500 images of 65 object classes in office and home settings. It has four dissimilar domains: Artistic images (\textbf{Ar}), ClipArt (\textbf{Cl}), Product images (\textbf{Pr}), and Real-World images (\textbf{Rw}). Among the four domains, there are a total of 12 DA tasks. 3) \noindent\textbf{VisDA-2017}~\cite{peng2017visda} is a simulation-to-real dataset for DA with over 280,000 images across 12 categories in the training, validation and testing domains. 
4) \noindent\textbf{DomainNet}~\cite{peng2019moment} is a benchmark for large-scale multi-source domain adaptation, which has six domains (Clipart, Infograph, Painting, Quickdraw, Real and Sketch)
and 0.6M images of 345 classes. 5) \noindent\textbf{Digit-Five} consists of five different digit recognition datasets: {MNIST (\textit{mt})~\cite{lecun1998gradient}, MNIST-M (\textit{mm})~\cite{ganin2015unsupervised}, USPS (\textit{up})~\cite{hull1994database}, SVHN (\textit{sv})~\cite{netzer2011svhn} , and Synthetic (\textit{syn})~\cite{ganin2015unsupervised}}. 


Following previous works {\cite{cui2020gradually,yang2020curriculum,jin2020feature}}, we perform single source to single target adaptation on the first three datasets.
\tcb{We also conduct experiments for multi-source domain adaptation on Digit-Five and DomainNet, following the leave-one-out protocol used in works~\cite{mancini2018boosting, peng2019moment}, where one domain is selected as the target domain while the rest of domains are used as source domains.}

\noindent\textbf{Implementation Details.}
\noindent\emph{Baseline Setup:} We build our baseline networks based on two representative domain adversarial domain adaptation frameworks: DANN~\cite{ganin2016domain} and CDAN~\cite{long2018conditional}. Following \cite{long2018conditional}, we could also use entropy conditioning (E) regularization as proposed in CDAN~\cite{long2018conditional}, which reweighs the samples based on their entropy of the \tcb{object} classification predictions to prioritize the easy-to-transfer samples for easy optimization. We denote the schemes with entropy conditioning (E) regularization as CDAN+E, DANN+E. We employ stochastic gradient descent (SGD) as optimizer with an initial learning rate of 1e-3 and momentum of 0.9 to train all the models. 




\noindent\emph{Our RADA Setup:} As a plug-and-play optimization strategy, we apply our RADA on top of the above two representative UDA baselines (DANN and CDAN) for validation. 
Batch size is set as 36 for Office-31 and VisDA-2017. More details are presented in \textbf{Supplementary}. As discussed in Section \ref{subsec:RADA}, we perform the ``well aligned" target sample selection, relabeling, and domain discriminator and feature extractor updating with the updated domain partitions at the mini-batch level. We train our RADA method for 100 epochs on Office-31 and Office-Home, and 150 epochs for the larger VisDA-2017 dataset. \tcb{We set threshould $\tau$ to \tcr{0.35} and $K$ to \tcr{5} by default, which we have investigated their influences in our ablation study section.}

\subsection{Ablation Study}~\label{sec:ablation_study}
\vspace{-4mm}

We perform comprehensive ablation studies to demonstrate the effectiveness of our proposed Re-enforceable Adversarial Domain Adaptation strategy. 
We conduct experiments on the two representative adversarial domain adaptation frameworks \emph{DANN}~\cite{ganin2016domain} and \emph{CDAN}~\cite{long2018conditional}.



\begin{table}[t]
  \centering
  \footnotesize
  \caption{Performance (classification accuracy \%) comparisons of our schemes and the corresponding baselines. ``w/o MU" denotes without using mixup within the updated source set. All schemes use ResNet-50 as backbone. 
  }
    \vspace{-2mm}
    \setlength{\tabcolsep}{1.8mm}{
    \begin{tabular}{c|ccc}
    \toprule
    Methods & Office-31 & Office-Home & VisDA-2017 \\
    \hline
    Baseline~(DANN~\cite{ganin2016domain}) & 83.42 & 60.05 & 61.23 \\
    \hline
          DANN+RADA w/o MU & 85.24 & 63.14 & 65.91 \\
          DANN+RADA  & \textbf{86.79} & \textbf{64.81} & \textbf{67.29} \\
    \hline
    \hline
     Baseline~(CDAN~\cite{long2018conditional}) & 87.90 & 68.11 & 70.82 \\
     \hline
          CDAN+RADA w/o MU & 89.58 & 70.25 & 75.62 \\
          CDAN+RADA  & \textbf{91.08} & \textbf{71.37}      & \textbf{76.28}  \\
    \bottomrule
    \end{tabular}}%
    \vspace{-4mm}
  \label{tab:ablation1}%
\end{table}%

\noindent\textbf{Effectiveness of RADA.}  Table~\ref{tab:ablation1} shows the averaged domain adaptive classification results on three benchmarks. We have the following observations:

\noindent\textbf{1)} On top of the baseline \emph{DANN}, \textbf{our \emph{DANN+RADA w/o MU}} even \emph{without using mixup} (for the updated source set) brings significant improvement of \textbf{1.82\%/3.09\%/4.68\%} in accuracy on Office-31/Office-Home/VisDA-2017, respectively. On top of the stronger baseline \emph{CDAN}, \textbf{our \emph{CDAN+RADA w/o MU}} still brings significant improvements, \ieno, \textbf{1.68\%/2.14\%/4.80\%} in accuracy on Office-31/Office-Home/VisDA-2017, respectively.  

\noindent\textbf{2)} Mixup within the updated source set promotes the continuity in the updated source set. It brings additional gain of {1.55\%/1.67\%/1.38\%} in accuracy on Office-31/Office-Home/VisDA-2017 on top of \emph{DANN+RADA w/o MU}. \tcb{Note that on VisDA-2017, the majority of gain comes from our relabeling strategy, \ieno, 4.68\%/4.80\% on DANN/CDAN, whereas the mixup specific to our updated source set brings additional gain of 1.38\%/0.66\%.}

\noindent\textbf{3)} The performance improvement of our \emph{RADA} over \emph{Baseline} on the large dataset VisDA-2017 is larger than that on the other two datasets, which reveals that our method is effective on the challenging dataset \tcb{with} large domain gap (synthetic$\rightarrow$real).

\begin{table}[htbp]
  \centering
  \footnotesize
  \caption{Comparison with re-weighting related methods for multi-source domain adaptation on digits classification. 
  }
  \vspace{-3mm}
  \setlength{\tabcolsep}{1.8mm}{
    \begin{tabular}{ccccccc}
    \toprule
    \multirow{2}[4]{*}{Methods} & \multicolumn{6}{c}{Digit-Five} \\
\cmidrule{2-7}          & \textit{mt}    & \textit{mm}    & \textit{sv}    & \textit{sy}    & \textit{up}    & \textit{Avg.} \\
    \midrule
    DANN~\cite{ganin2014unsupervised}  & 97.9  & 70.8  & 68.5  & 87.3  & 93.4  & 83.6 \\
    DANN + E~\cite{long2018conditional} & 98.4  & 73.7  & 71.5  & 87.8  & 94.7  & 85.2 \\
    
    DANN + inverse-E~\cite{long2018conditional} & 98.7  & 74.4  & 72.4  & 88.6  & 96.4  & 86.1 \\

    DANN+IWAN~\cite{zhang2018importance}  & 98.2  & 74.2  & 72.9  & 88.9  & 95.8  & 86.0 \\
    DANN+RADA w/o MU  & 99.0  & 76.6  & 88.9  & 94.7  & 98.4  & 91.5 \\
    DANN+RADA w/ MU  & \textbf{99.3}  & \textbf{78.2}  & \textbf{90.0}  & \textbf{95.2}  & \textbf{98.4}  & \textbf{92.2} \\   
    \bottomrule
    \end{tabular}}%
    \vspace{-5mm}
  \label{tab:digits_result}%
\end{table}%

\noindent\textbf{Comparison with Sample Re-weighting based Methods.} 
We compare RADA with some sample (re)weighting based schemes, including entropy-based re-weighting (E) ~ \cite{long2018conditional}, IWAN~\cite{zhang2018importance}.
Entropy-based re-weighting (E) aims to prioritize the easy-to-transfer samples to ease the optimization\tcr{, sharing a similar idea in curriculum learning}. IWAN~\cite{zhang2018importance} re-weights the samples based on the outputs of domain discriminator, which assigns small weights to those source samples that are easy to be distinguished by the domain discriminator to exclude the outlier classes in the source domain. 
\emph{DANN+inverse-E} denotes the scheme that after half of the optimization iterations, we turn the focus (large weight) from easy-to-transfer samples to those hard-to-aligned samples to enable hard-mining, by adjusting entropy-based reweighting strategy from  $\omega(ent(\cdot))=1+e^{-ent(\cdot)}$ to $\omega(ent(\cdot))=1 / (1+e^{-ent(\cdot)})$, where $ent(\cdot)$ denotes the entropy of the predicted object classes.

We conduct experiments under the multi-source setup on Digit-Five. To guarantee fairness of comparison, all competitors use the same network architecture (the feature extractor is composed of three convolutional layers and two FC layers ($Cov_{3}FC_{2}$)), and are built upon the representative domain adversarial training framework DANN~\cite{ganin2016domain}.


Table~\ref{tab:digits_result} shows the results. We can see that all sample re-weighting strategies bring performance gain. Our RADA strategy outperforms the performance of all these strategies. \tcb{In essence, the re-weighting based methods adjust the levels of importance of samples during optimization (from the perspective of curriculum learning or hard mining). They all use static domain labels in optimization. 
Giving larger weights to the ``well aligned" samples actually provides more aligned distributions to the domain discriminator, which cannot effectively enhance the discrimination capability of the domain classifier.  
Giving smaller weights to the ``well aligned" samples may result in less effective use of them for training the domain discriminator. 
\emph{In contrast, we allow dynamic domain labels by relabeling those ``well aligned" target domain samples as source domain samples. On one hand, this provides more separable distributions (vs. more aligned distributions as in assigning larger weights) to re-energize the domain discriminator. On the other hand, this enables the full use of ``well aligned" samples (vs. less effective use as in assigning smaller weights).}}

\subsection{Design Choices}~\label{sec:choices}
\vspace{-4mm}

For clear analysis, we do experiments on our scheme \emph{RADA w/o MU} (\ieno, without using mixup for the updated source set) for design choices study on VisDA-2017.

\begin{figure}
  \centerline{\includegraphics[width=1.0\linewidth]{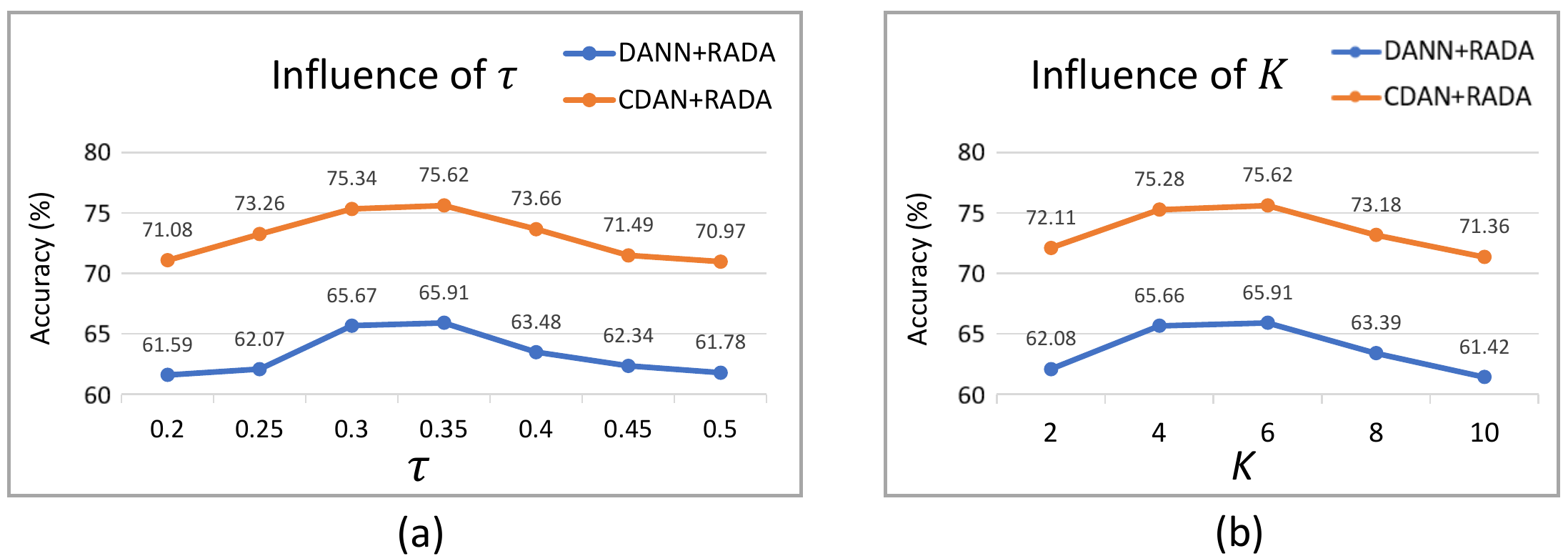}}
  \vspace{-2mm}
  \caption{Influence of (a) threshold $\tau$, and (b) $K$.}
  \vspace{-3mm}
  \label{fig:global_entropy}
\end{figure}

\noindent\textbf{Influence of Threshold $\tau$.}
\tcb{As described in Section~\ref{subsec:RADA}, we employ a hyper-parameter $\tau$ as the threshold to determine whether a target sample is ``well aligned". We study its influence in Figure~\ref{fig:global_entropy}~(a). We can see that a superior performance is achieved when $\tau$ ranges from 0.3 to 0.35 on two schemes. Similar trends are observed on other datasets. We set $\tau=0.35$ on all datasets. Note that we set $K=5$ for all these comparison experiments.}

\noindent\textbf{Influence of $K$.} \tcb{As described in Section \ref{subsec:RADA}, we start our RADA if no improvement of the discrimination capability of the domain classifier is seen for $K$ number of epochs.  Figure~\ref{fig:global_entropy}~(b) shows the influence of $K$. We observe that superior performance is achieved when $K$ ranges from 4 to 6. Similar trends are observed on other datasets. We set $K=5$ for all datasets. When $K$ is too small, the judgement of no improvement is not reliable which is sensitive to noise. When $K$ is too large, this optimization strategy cannot be fully utilized.}

\begin{figure}
  \centerline{\includegraphics[width=0.95\linewidth]{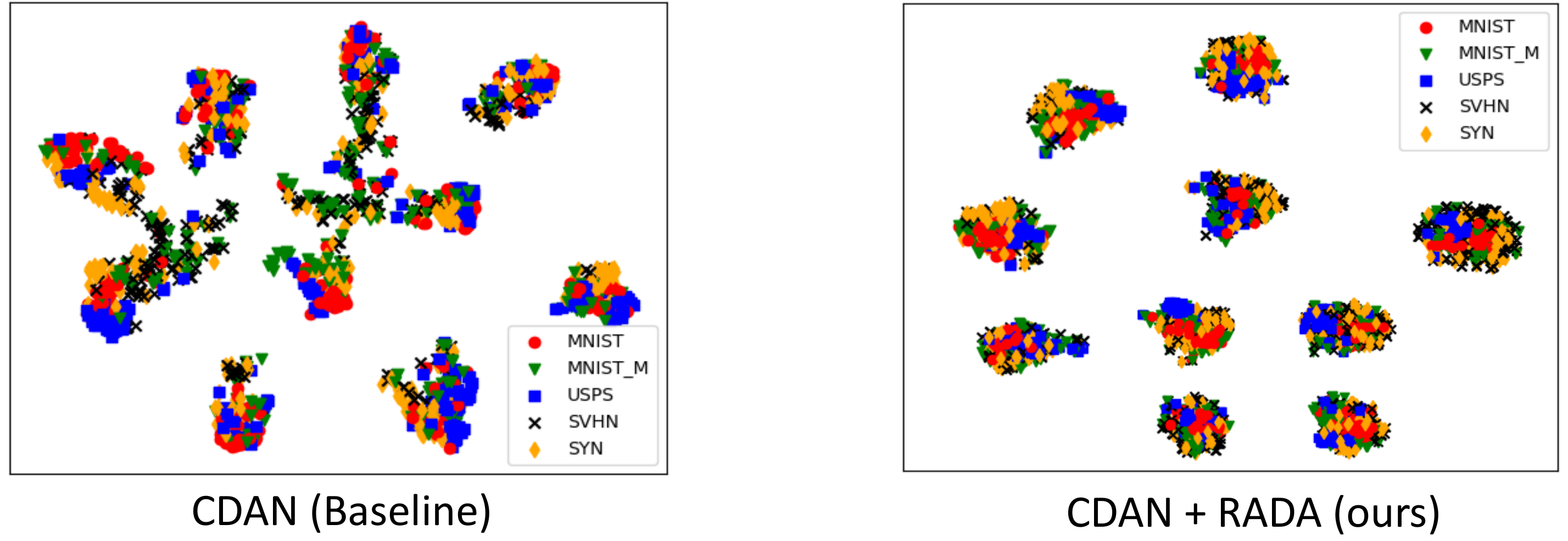}}
  \vspace{-2mm}
  \caption{Visualization of t-SNE distributions on Digit-Five.
  Compared with CDAN, the features of different domains are better aligned and the different classes/digits are clearly separated for our CDAN+RADA.}
  \vspace{-4mm}
  \label{fig:vis_tSNE_common}
\end{figure}

\begin{table*}[]
  \centering
  \small
  \caption{Performance (\%) comparisons with the state-of-the-art UDA approaches on Office-31. All experiments are conducted based on ResNet-50 pre-trained on ImageNet.}
  \vspace{-2mm}
    \begin{tabular}{ccccccccc}
    \toprule
    Method & Venue & A$\rightarrow$D  & A$\rightarrow$W  & D$\rightarrow$W  & W$\rightarrow$D  & D$\rightarrow$A  & W$\rightarrow$A  & Avg. \\
    \midrule
    DANN~\cite{ganin2016domain} & JMLR'16 & 79.7$\pm$0.4 & 82.0$\pm$0.4 & 96.9$\pm$0.2 & 99.1$\pm$0.1 & 68.2$\pm$0.4 & 67.4$\pm$0.5 & 82.2 \\
    Simnet~\cite{pinheiro2018unsupervised} & CVPR'18 & 85.3$\pm$0.3 & 88.6$\pm$0.5 & 98.2$\pm$0.2 & 99.7$\pm$0.2 & 73.4$\pm$0.8 & 71.8$\pm$0.6 & 86.2 \\
    MCD~\cite{saito2018maximum} & CVPR'18 & 92.2$\pm$0.2 & 88.6$\pm$0.2 & 98.5$\pm$0.1 & \textbf{100.0$\pm$.0} & 69.5$\pm$0.1 & 69.7$\pm$0.3 & 86.5 \\
    CDAN~\cite{long2018conditional} & NIPS'18 & 92.9$\pm$0.2 & 93.1$\pm$0.1 & 98.6$\pm$0.1 & \textbf{100.0$\pm$.0} & 71.0$\pm$0.3 & 69.3$\pm$0.3 & 87.5 \\
    TADA~\cite{wang2019transferable} & AAAI'19 & 91.6$\pm$0.3 & 94.3$\pm$0.3 & 98.7$\pm$0.1 & 99.8$\pm$0.2 & 72.9$\pm$0.2 & 73.0$\pm$0.3 & 88.4 \\
    Symnets~\cite{zhang2019domain} & CVPR'19 & 93.9$\pm$0.5 & 90.8$\pm$0.1 & 98.8$\pm$0.3 & \textbf{100.0$\pm$.0} & 74.6$\pm$0.6 & 72.5$\pm$0.5 & 88.4 \\
    CAN~\cite{kang2019contrastive} & CVPR'19 & 95.0$\pm$0.3 & 94.5$\pm$0.3 & 99.1$\pm$0.2 & 99.8$\pm$0.2 & \textbf{78.0$\pm$0.3} & 77.0$\pm$0.3 & 90.6 \\
    GVB~\cite{cui2020gradually} & CVPR'20 & 95.0$\pm$0.4 & 94.8$\pm$0.5 & 98.7$\pm$0.3 & \textbf{100.0$\pm$.0} & 73.4$\pm$0.3 & 73.7$\pm$0.4 & 89.3 \\
    CDAN-GVB~\cite{cui2020gradually} & CVPR'20 & 93.7$\pm$0.2 & 94.0$\pm$0.2 & 98.6$\pm$0.1 & \textbf{100.0$\pm$.0} & 73.4$\pm$0.3 & 73.0$\pm$0.2 & 88.8 \\
    SRDC~\cite{tang2020unsupervised} & CVPR'20 & 95.8$\pm$0.2 & 95.7$\pm$0.2 & 99.2$\pm$0.1 & \textbf{100.0$\pm$.0} & 76.7$\pm$0.3 & 77.1$\pm$0.1 & 90.8 \\
    \midrule
    CDAN (Baseline)~\cite{long2018conditional} & NIPS'18 & 90.8$\pm$0.3 & 94.0$\pm$0.5 & 98.1$\pm$0.3 & \textbf{100.0$\pm$.0} & 72.4$\pm$0.4 & 72.1$\pm$0.3 & 87.9 \\
    CDAN+RADA & This work & \textbf{96.1$\pm$0.4} & \textbf{96.2$\pm$0.4} & \textbf{99.3$\pm$0.1} & \textbf{100.0$\pm$.0} & 77.5$\pm$0.1 & \textbf{77.4$\pm$0.3} & \textbf{91.1} \\
    \bottomrule
    \end{tabular}%
  \label{tab:office31}%
\end{table*}%

\begin{table*}[]\centering
\vspace{-1mm} 
    \caption{Performance (\%) comparisons with the state-of-the-art UDA approaches on Office-Home, VisDA-2017, Digit-Five, and DomainNet. For all these approaches, ResNet-50 is taken as backbone for Office-Home, VisDA-2017,  $Cov_{3}FC_{2}$ is taken as backbone for Digit-Five, ResNet-101 is taken as backbone for DomainNet. See \textbf{Supplementary} for more results.}
    \vspace{-2mm}
	\captionsetup[subfloat]{captionskip=2pt}
	\captionsetup[subffloat]{justification=centering}
	\subfloat[Comparison on Office-Home.\label{tab:office_home}]{
		\tablestyle{1.8pt}{0.94}
        \begin{tabular}{c|c|c}
        \toprule
        Method & Venue & Avg. \\
        \midrule
        DANN~\cite{ganin2016domain} & JMLR'16 & 57.6 \\
        MCD~\cite{saito2018maximum} & CVPR'18 & 64.1 \\
        CDAN~\cite{long2018conditional} & NIPS'18 & 65.8 \\
        Symnets~\cite{zhang2019domain} & CVPR'19 & 67.2 \\
        TADA~\cite{wang2019transferable} & AAAI'19 & 67.6 \\
        BNM~\cite{cui2020nnm} & CVPR'20 & 67.9 \\
        CDAN-GVB~\cite{cui2020gradually} & CVPR'20 & 69.0 \\
        GVB~\cite{cui2020gradually} & CVPR'20 & 70.4 \\
        SRDC~\cite{tang2020unsupervised} & CVPR'20 & 71.3 \\
        \midrule
        CDAN (Baseline) & NIPS’18 & 68.1 \\
        CDAN+RADA & This work & \textbf{71.4} \\
        \bottomrule
        \end{tabular}%
    }
    \hspace{0.5mm}
	\subfloat[Comparison on VisDA-2017. \label{tab:VisDA_2017}]{
		\tablestyle{1.8pt}{1.12}
		    \begin{tabular}{c|c|c}
            \toprule
            Method & Venue & Avg. \\
            \midrule
            DAN~\cite{long2015learning} & ICML'15 & 61.6 \\
            DANN~\cite{ganin2016domain} & JMLR'16 & 57.4 \\
            GTA~\cite{sankaranarayanan2018generate} & CVPR'18 & 69.5 \\
            CDAN~\cite{long2018conditional} & NIPS'18 & 70.0 \\
            MDD~\cite{zhang2019bridging} & ICML'19 & 74.6 \\
            CDAN-GVB~\cite{cui2020gradually} & CVPR'20 & 74.9 \\
            GVB~\cite{cui2020gradually} & CVPR'20 & 75.3 \\
            \midrule
            CDAN (Baseline) & NIPS’18 & 70.8 \\
            CDAN+RADA & This work & \textbf{76.3} \\
            \bottomrule
            \end{tabular}%
                }
                \hspace{0.5mm}
	\subfloat[Comparison on Digit-Five.\label{tab:digit5}]{
		\tablestyle{1.8pt}{1.12}
		\begin{tabular}{c|c|c}
        \toprule
        Methods & Venue & Avg. \\
        \midrule
        DANN~\cite{ganin2016domain} & JMLR'16 & 83.6 \\
        IWAN~\cite{you2019towards} & ICML'19 & 86.0 \\
        MDAN~\cite{zhao2018adversarial} & NIPS'18 & 86.7 \\
        MCD~\cite{saito2018maximum} & CVPR'18 & 86.1 \\
        DCTN \cite{xu2018deep} & CVPR'18 & 88.6 \\
        M3SDA~\cite{peng2019moment} & ICCV'19 & 87.6 \\
        CMSS~\cite{yang2020curriculum} & ECCV'20 & 90.8 \\
        \midrule
        CDAN (Baseline) & NIPS’18 & 88.7 \\
        CDAN+RADA & This work & \textbf{93.2} \\
        \bottomrule
        \end{tabular}%
	}
	\hspace{0.5mm}
	\subfloat[Comparison on DomainNet. \label{tab:DomainNet}]{
		\tablestyle{1.8pt}{1.12}
		    \begin{tabular}{c|c|c}
            \toprule
            Methods & Venue & Avg. \\
            \midrule
            DANN~\cite{ganin2016domain} & JMLR'16 & 32.7 \\
            DCTN \cite{xu2018deep} & CVPR'18 & 38.3 \\
            MCD~\cite{saito2018maximum} & CVPR'18 & 38.5 \\
            MDAN~\cite{zhao2018adversarial} & NIPS'18 & 42.8 \\
            M3SDA~\cite{peng2019moment} & ICCV'19 & 42.7 \\
            FAR~\cite{jin2020feature} & Arxiv'19 & 45.5 \\
            CMSS~\cite{yang2020curriculum} & ECCV'20 & 46.5 \\
            \midrule
            CDAN (Baseline) & NIPS’18 & 45.2 \\
            CDAN+RADA & This work & \textbf{47.5} \\
            \bottomrule
            \end{tabular}%
                }
	\vspace{-8mm}
	\label{tab:officehome_visDA_digit5_domainNet}
\end{table*}

\subsection{Visualization}~\label{sec:Vis}

\vspace{-4mm}

\noindent\textbf{Visualization of Feature Distributions.} 
In Figure \ref{fig:vis_tSNE_common}, we visualize the distributions of the features using t-SNE \cite{van2008visualizing} in \emph{mm,mt,sv,syn$\rightarrow$up} setting on Digit-Five. We compare the feature distribution with the baseline CDAN~\cite{long2018conditional}, and observe that the features of different domains are better aligned and the different classes/digits are clearly separated for our scheme CDAN+RADA.

\subsection{Comparison with State-of-the-Arts}~\label{sec:SOTA}
\vspace{-3mm}

We compare our scheme CDAN~\cite{long2018conditional}+RADA with state-of-the-art unsupervised single/multi-source domain adaptation methods. We report the results from their original papers if available. For fair comparison, we also report the result of our baseline scheme CDAN run by us. 
\tcb{SRDC~\cite{tang2020unsupervised}, CAN~\cite{kang2019contrastive}, and Symnets~\cite{zhang2019domain} tend to encourage the category/group-level consistency of distributions to alleviate the damage of intrinsic discrimination of data when minimizing the domain discrepancy. GVB~\cite{cui2020gradually} and CMSS~\cite{yang2020curriculum} explore to reduce domain discrepancy in a curriculum manner. Ours is conceptually complementary to these methods. They in general ignore the degradation of the discrimination capability of the domain discriminator and its side effect. To address this, we propose RADA which could re-energize the domain discriminator by exploring dynamic domain labels, without making any change to the network architecture of the domain discriminator.}

\noindent\textbf{Results on Office-31, Office-Home and VisDA-2017.} For the single source to single target adaptation, Table~\ref{tab:office31}, Table~\ref{tab:office_home} and Table~\ref{tab:VisDA_2017} show the comparisons on Office-31, Office-Home, and VisDA-2017. We can see that our CDAN+RADA significantly outperforms the baseline CDAN~\cite{long2018conditional} by \textbf{3.2\%, 3.3\%, and 5.5\%} on three datasets. CDAN+RADA also achieves the best performance. 



\noindent\textbf{Results on Digit-Five and DomainNet.} We also conduct experiments on the multi-source to single target adaptation settings on Digit-Five and DomainNet. On Table \ref{tab:digit5} and Table~\ref{tab:DomainNet} show that our CDAN+RADA  outpeforms our baseline CDAN by 4.5\% and 2.3\% on Digit-Five and DomainNet, respectively. Our CDAN+RADA also achieves the state-of-the-art performance. 
More experimental results, including various sub-settings on different single/multi-source UDA datasets, can be found in \textbf{Supplementary}.


\vspace{-2mm}
\section{Conclusion}
\vspace{-1mm}

In this paper, we propose an effective optimization strategy for adversarial domain adaptation, termed as Re-enforceable Adversarial Domain  Adaptation (RADA), which aims to re-energize the domain classifier during the training and in turn further drives feature alignment. We achieve this by relabeling the ``well aligned'' target samples as source domain\tcb{, which enables the exploration of dynamic domain labels,} for re-energizing the domain classifier on the fly. Extensive experiments on multiple benchmarks, including single- and multi-source domain adaptation, and different adversarial domain adaptation networks, have demonstrated the effectiveness and generalization capability of our RADA strategy for adversarial domain adaptation. 
\tcb{We expect this work to inspire more works which investigate/incorporate more dynamics to the adversarial domain adaption for effective training.}



{\small
\bibliographystyle{ieee_fullname}
\bibliography{egbib}
}

\end{document}